\documentclass{article}
\usepackage{spconf,amsmath,epsfig}

\let\OLDthebibliography\thebibliography
\renewcommand\thebibliography[1]{
  \OLDthebibliography{#1}
  \setlength{\parskip}{0pt}
  \setlength{\itemsep}{0pt plus 0.3ex}
}

\usepackage{algorithm}
\usepackage[noend]{algorithmic}
\usepackage{amsmath}
\usepackage{amssymb}
\usepackage{booktabs}
\usepackage{subfigure}
 \usepackage{multirow}

\begin{document}\sloppy

% Example definitions.
% --------------------
\def\x{{\mathbf x}}
\def\L{{\cal L}}

% Title.
% ------
\title{Generative Steganographic Flow}
%
% Single address.
% ---------------
\name{Ping Wei, Ge Luo, Qi Song, Xinpeng Zhang*, Zhenxing Qian*, Sheng Li}

%Address and e-mail should NOT be added in the submission paper. They should be present only in the camera ready paper. 
\address{School of Computer Science, Fudan University, Shanghai, China \\ \{pwei17, 18110240026, 19210240114, zhangxinpeng, zxqian, lisheng\}@fudan.edu.cn}

\maketitle

\footnotetext[1]{The corresponding authors are Xinpeng Zhang and Zhenxing Qian.}

\begin{abstract}
Generative steganography (GS) is a new data hiding manner, featuring direct generation of stego media from secret data. Existing GS methods are generally criticized for their poor performances. In this paper, we propose a novel flow based GS approach -- Generative Steganographic Flow (GSF), which provides direct generation of stego images without cover image. We take the stego image generation and secret data recovery process as an invertible transformation, and build a reversible bijective mapping between input secret data and generated stego images. In the forward mapping, secret data is hidden in the input latent of Glow model to generate stego images. By reversing the mapping, hidden data can be extracted exactly from generated stego images. Furthermore, we propose a novel latent optimization strategy to improve the fidelity of stego images. Experimental results show our proposed GSF has far better performances than SOTA works.
\end{abstract}
\section{Introduction}
Steganography usually  modifies cover media slightly to hide secret data\cite{fridrich2009steganography}.  Earlier image-based steganographic algorithms usually hide secret data in the least significant bits of cover images. Later, lots of syndrome trellis coding (STC) based algorithms are put forward to minimize the distortion after data embedding\cite{survey_1}.  As deep learning (DL) develops, many DL-based steganographic solutions have also been proposed\cite{survey_2}. For example,  Baluja\cite{img_in_img} presents a neural network to hide secret images in a cover image. However, there are non-negligible differences between the generated stego images and cover images in these methods, which can be detected easily by steganalysis algorithms\cite{color_rich, srnet}. Where, Steganalysis is the counterpart technique to detect whether images contain secret data. 
\begin{figure}[t]
	\centering
	\includegraphics[width=3 in]{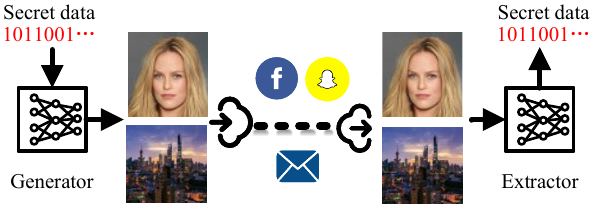} 
	\caption{The flowchart of generative steganography.  Binary secret data is converted to stego images with an image generator. Then, the generated stego images can be transmitted through social networks, email or other channels. At last, the hidden secret data is recovered from received stego images using the data extractor.}
	\label{Fig.1}
\end{figure} 

Recently, a new steganographic manner called generative steganography (GS) has attracted researchers’ attention\cite{coverless}. With this approach, secret data is directly converted to natural stego images, i.e., cover images are no longer needed. The flowchart of GS is illustrated in Fig.\ref{Fig.1}. In the early stage, people used to synthesize some particular types of image to conceal secrets, like texture image \cite{wkc} and fingerprint image\cite{lisheng}. However, only some  special  image types can be created by this kind of methods, and their unnatural contents may arouse suspicion of the monitor. Later,  some researchers\cite{zhang2020generative,hu2018novel} proposed to synthesize natural stego images with generative adversarial networks (GANs). As for these solutions, the input variables of GANs are often used to represent secret data by a pre-built mapping dictionary. Natural stego images are generated with GANs by assigning the corresponding labels or noise vectors. However, their performances are poor with low image quality and hiding capacity. 

In this paper, we propose a novel flow-based GS method called GSF. Unlike existing GS methods that require a network to generate stego images and the other one to extract the hidden secrets,  our proposed GSF accomplishes both tasks on one network. Instead of using mapping dictionary as most GS methods, we propose to hide secret data in the binarized latent of Glow for high performance. An invertible bijective mapping is built for the lossless transformation between secret data and stego images. Our main contributions are:

\begin{itemize}
	\item We first propose a flow-based GS solution that can generate realistic stego images and recover the hidden secret using a single network.
	\item We build a reversible bijective mapping between the input secret data and generated stego images, with which secret data and stego images can convert to each other almost without information loss. 
	\item  A novel latent optimization strategy is developed to improve the quality of stego images.
	\item  Our proposed GSF achieves superior performances than existing works. The hiding capacity, extraction accuracy, stego image quality and security of our scheme are much higher than those of state-of-the-art works.  
\end{itemize}

\section{Generative Steganographic Flow}
\begin{figure}[t]
	\centering
	\includegraphics[width=\linewidth]{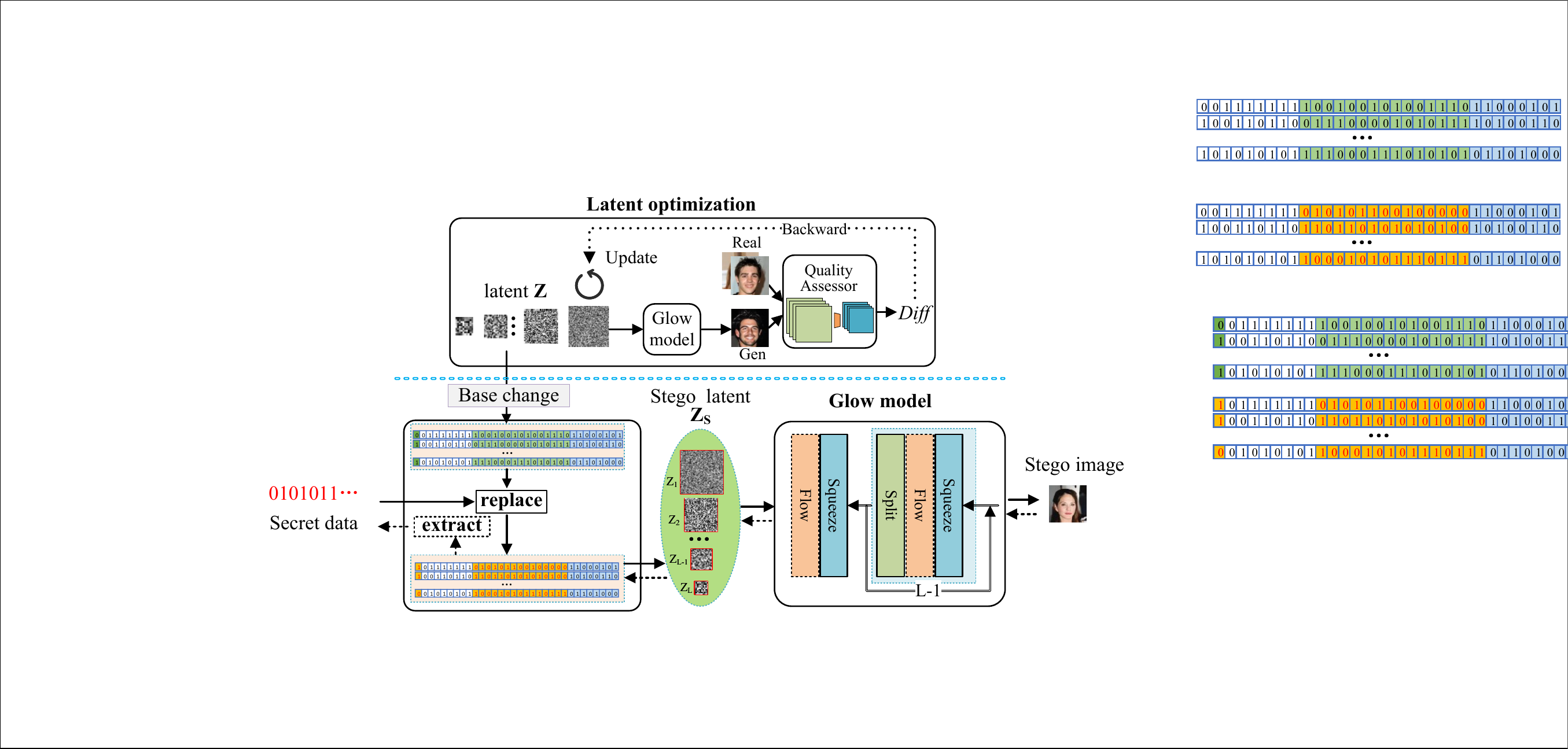} 
	\caption{the overall  structure of our proposed GSF. Latent \textbf{Z} is updated circularly with the backward loss \textit{Diff}, which is calculated by an image quality assessor. Secret data is hidden in the binarized sequences of optimized \textbf{Z}. Next, the modified sequences are converted to $\mathbf{Z_s}$ and then sent to Glow model for stego image generation. Hidden secret data can be extracted by reversing the hiding process.}
	\label{Fig.total}
\end{figure}

The pipeline of proposed GSF is described in Fig.\ref{Fig.total}, which consists of two stages: 1. the stage of latent optimization; 2. the stage of secret hiding and extracting. In the first stage, the latent \textbf{Z} is optimized,  aiming at improving the quality of generated stego images. As shown on the top of Fig.\ref{Fig.total}, a quality assessor is adopted to assess the image fidelity, with which, a score will be outputted depending on the fidelity of every input image. The score difference (\textit{Diff}) between the generated image and real images is used as the backward loss for optimization; In the second stage, a reversible bijective mapping is built from the input secret data to the output stego images, by which secret data can be converted to a stego image and vice versa. The data hiding process is described at the bottom of Fig.\ref{Fig.total}. Firstly, the optimized latent \textbf{Z} in the first stage is encoded to binary sequences, and some hideable bits in each of these sequences are replaced by the input secret data. Secondly, these modified sequences are converted to \textit{L} float-pointed tensors that share the same size with tensors in \textbf{Z}, named stego latent $ \mathbf{Z_s} $. Finally, $ \mathbf{Z_s} $ is sent to a pre-trained Glow model for generating stego images. Inversely, the hidden secret data can be retrieved by reversing the procedures above.

\subsection{Model training}
Our scheme is built on the basis of the flow-based generative model Glow\cite{glow}, with which, the input latent \textbf{Z} and the generated image \textbf{I} can convert to each other almost without information loss, i.e., $Glow(\mathbf{I})=\mathbf{Z}$ and $Glow^{-1}(\mathbf{Z})=\mathbf{I}$. $Glow$ is composed of a series of invertible functions,  $Glow=f_1 \times f_2 \cdots \times f_n$. The  transformation of Glow can be expressed as:
$
\mathbf{I} \stackrel{f_{1}}{\rightarrow} \mathbf{h_{1}} \stackrel{f_{2}}{\rightarrow} \mathbf{h_2} \cdots \stackrel{f_{n}}{\rightarrow} \mathbf{Z}, 		\mathbf{Z} \stackrel{f_{n}^{-1}}{\longrightarrow} \mathbf{h_{n-1}}  \cdots \stackrel{f_{2}^{-1}}{\longrightarrow} \mathbf{h_1} \stackrel{f_{1}^{-1}}{\longrightarrow} \mathbf{I} ,
$
here, $ f_i $ is a reversible transformation function, and $ \mathbf{h_i} $ is the output of $ f_i $. 
Glow is composed of three types of modules, including the squeeze module, the flow module and the split module. The squeeze module is used to downsample the feature maps, and the flow module is used for feature processing.  The split module will divide the image features into halves along the channel side, and half of them are outputted as the latent tensor $ \mathbf{Z_i} $, which has the size of: $(h_{i}, w_{i},  c_{i})=\left\{\begin{array}{l}(\frac{H}{2^{i}}, \frac{W}{2^{i}}, 6 \times 2^{i}), ~ i=0, \cdots, L-2 \\ (\frac{H}{2^{i}}, \frac{W}{2^{i}}, 12 \times 2^{i}), \quad \quad ~ i=L-1\end{array}\right.$, where $H$/$W$ are the height/width of generated images,  and $L$ is the number of $ \mathbf{Z_i}$. Then, the other half of features are cycled into the squeeze module. In our scheme, $\mathbf{Z}=\sum_{i=1}^{L} \mathbf{Z_{i}}$  is regarded as the overall input latent of Glow.
The loss function of the Glow model is defined as: 
\begin{equation}
	\label{eq-1}
	\begin{aligned}
		\log p_{\boldsymbol{\theta}}(\mathbf{I}) =\log p_{\boldsymbol{\theta}}(\mathbf{Z})+\sum_{i=1}^{n-1} \log \left|\operatorname{det}\left(d \mathbf{h}_{i} / d \mathbf{h}_{i-1}\right)\right| ,
	\end{aligned}
\end{equation}
where,  $\log |\operatorname{det}\left(d \mathbf{h}_{i} / d \mathbf{h}_{i-1}\right)|$ is the logarithm of the absolute value of the determinant of the Jacobian matrix $d \mathbf{h}_{i} / d \mathbf{h}_{i-1}$.

\subsection{Latent Optimization strategy}
The initial input latent of Glow follows the distribution: $ \mathbf{Z} \sim N (0, 1) \times \delta $, however, it can only generate low-quality images.  Therefore, we propose a latent optimization strategy to improve the quality of generated images.  The proposed strategy is depicted on the top of Fig.\ref{Fig.total} and described in algorithm 1. Instead of using a random normal distribution, we initialize \textbf{Z} with the average value of 
projected latents, where \textit{n} randomly sampled real images $\mathbf{I_n}$ are converted to latent values using Glow model, i.e., $\mathbf{Z}=\frac{1}{n} \sum Glow (\mathbf{I_n})$. Next, the initialized \textbf{Z} is updated circularly with the backward loss $ Diff $ to further improve the image quality. 

\begin{algorithm}[ht]
	\caption{~ Latent optimization strategy}
	\label{alg:1}
	\textbf{Input}: \textit{n} randomly sampled real images $ \mathbf{I_n} $.\\
	\textbf{Output}: optimized latent \textbf{Z}. \\
	\textbf{Required}: A pre-trained Glow model, and a pre-trained Resnet50 as quality assessor (QA).
	\begin{algorithmic}[1] %[1] enables line numbers
		\STATE Initialize the latent, $\mathbf{Z}=\frac{1}{n} \sum Glow (\mathbf{I_n})$.
		\STATE QA outputs a score for the $i_{th}$ image of $ \mathbf{I_n} $, $score_{real}^i$.
		\FOR { \textit{step} in range (\textit{max-step}) }
		\STATE Glow generates an image (\textbf{Gen}) with \textbf{Z}.
		\STATE QA outputs a score for \textbf{Gen},  $score_{gen}$. 
		\STATE Calculate  $ Diff = \mid\frac{1}{n} \sum_{1}^{n} score_{real}^i - score_{gen}\mid$.
		\STATE Update \textbf{Z} with gradient $ \nabla_{\mathbf{z}} Diff $.
		\IF {\textit{Diff} $\textless $ \textit{thresh}}
		\STATE	Save the optimized latent \textbf{Z}.
		\STATE	Break.
		\ENDIF
		\ENDFOR
	\end{algorithmic}
\end{algorithm}

In the strategy, we use a quality assessor (QA) to evaluate the fidelity of images. A pre-trained classifier Resnet50\cite{resnet} is used as the QA, which is trained on real images and the generated images of Glow. The QA will output positive scores for real input images and negative scores for generated images.  $Diff$ is the difference between QA's outputted mean score for \textit{n} real images and the score for the generated image, which can be calculated as: 
\begin{equation}
	Diff = \mid \frac{1}{n} \sum_{1}^{n} score_{real}^i - score_{gen} \mid ,
	\label{Eq.3}
\end{equation}
where, $ score_{real}^i $ is  QA's outputted score for the $i_{th}$ real image, and $ score_{gen} $ is the score for the generated image. Only one image is generated in each step.
In each optimizing step, \textbf{Z} is updated as follows:
\begin{equation}
	\mathbf{Z}:=\mathbf{Z}-\varepsilon \cdot \nabla_{\mathbf{z}} Diff ,
	\label{Eq.4}
\end{equation}
here, $\varepsilon$ is a hyper-parameter controlling the change level, and $ \nabla_{\mathbf{z}} Diff $ is the gradient of latent \textbf{Z}. The optimization process will continue until the maximum training step is reached or the value of $ Diff $ is lower than the threshold.

\subsection{Secret Hiding and Extracting}

\begin{figure}[htb]
	\raggedleft
	\includegraphics[width=0.9\linewidth]{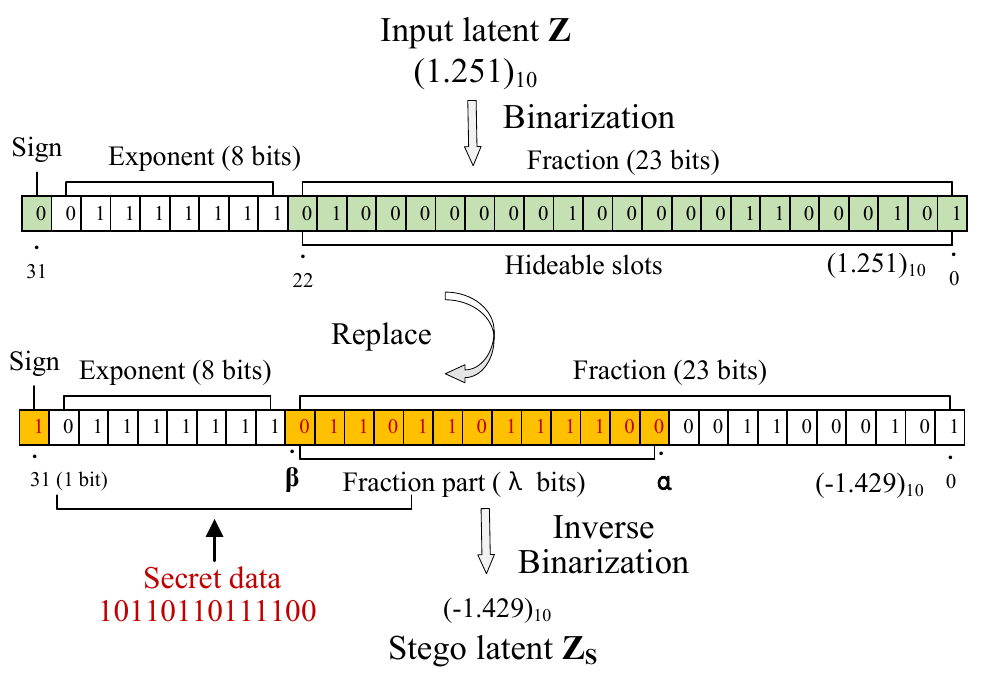} 
	\caption{schematic diagram of secret hiding. After converting \textbf{Z} to binarized sequences, input secret data is hidden in the hideable bits of these sequences. Then, the modified sequences are converted to float-point tensors named $\mathbf{Z_s}$. }
	\label{Fig.replace}
\end{figure}

Secret data is hidden in the binarized sequences of optimized latent as shown in Fig.\ref{Fig.total} and Fig.\ref{Fig.replace}. The IEEE 754 standard is used for base change, by which, each float-point number of \textbf{Z} is encoded to a 32-bit binary sequence by:
\begin{equation}
	(n)_{10}=(-1)^{sign } \times 2^{(exponent)_{10}-127} \times\left(1+(fraction)_{10}\right),
	\label{Eq.tran}
\end{equation}
here, $ (n)_{10} $ means a decimal float-point number. $ sign $  is Sign bit with the value of 0 or 1.  $ (exponent)_{10} $ indicates the decimal exponential multiplier, which is often equal to 125, 126 or 127 in our scheme. $ (fraction)_{10} $ denotes the fractional part of the float-point number. Then, the $ sign$ bit, the $ (exponent)_{10} $ and $ (fraction)_{10} $ values are converted to binary digits successively, forming a 32-bit binary sequence. 

The detailed data hiding process is demonstrated in Fig.\ref{Fig.replace}. First, all the float-point numbers of \textbf{Z} are encoded to 32-bit binary sequences with the function defined in Eq.\ref{Eq.tran}.  Except for the exponential bits that affect the image quality greatly, the $ Sign $ bit and the fractional part ( $ 0_{th} $ to $ 22_{th} $ bits) of  these binarized sequences can be used to conceal secret data, which are called the hideable bits.
Among them, only the  $ \alpha_{th} $ to $ \beta_{th} $ ( $ 0 \le \alpha \le \beta \le 22 $, $[\alpha:\beta] $) bits of the fraction part are employed for data hiding.  In our scheme,  we change the value of $\alpha $ to convey different secret payloads with $\beta$  fixed to 22. Then the modified sequences containing secret data are converted to float-point numbers again and then reconstituted to stego latent $ \mathbf{Z_{s}}$.

We can inverse the data hiding process to extract secret data. Firstly, the received stego image is sent to Glow model to regain the stego latent. Then, the recovered stego latent $\mathbf{Z_s^{*}} $ is converted to binary sequences. Finally, the hidden secret data can be retrieved from the $ Sign $ bit and the $[\alpha:\beta] $ bits of these sequences.
In summary, the  hiding and extracting process of secret data can be described as:
\begin{equation}
	\begin{split}
		&\mathbf{Secret} \rightarrow \mathbf{bin}(\mathbf{Z})[S, \alpha:\beta], \\  
		& \mathbf{Secret^{*}} \leftarrow \mathbf{bin}(\mathbf{Z_s^{*}})[S, \alpha:\beta],
	\end{split}
\end{equation}
where, $\mathbf{bin}(\mathbf{\cdot}) $ means converting the float-point numbers in latent tensors into binary sequences.  $\mathbf{Secret} $ is the input secret data, which is hidden in the $ Sign $ bit and the $\alpha$  - $\beta$ bits of  $ \mathbf{bin}(\mathbf{Z}) $, i.e.,  $ [S, \alpha:\beta] $.  $ \mathbf{Secret^{*}} $ is the extracted secret data retrieved from the same positions of $ \mathbf{bin}(\mathbf{Z_s^{*}})$.

\section{Experimental results} 
Our scheme is implemented with PyTorch on CentOS 7 with 4 Nvidia 1080Ti. CelebA\cite{celeba} is used to evaluate the performance of our model.  Adam is the optimizer, with a learning rate of 1e-3. The Glow model is trained to generate $ 128\times128 $ images with  $L=5 $. Parameter $\varepsilon$  in Eq.\ref{Eq.4} is set to 1e-3, and \textit{n} in Eq.\ref{Eq.3} and algorithm 1 is set to 3. $ thresh$  in algorithm 1 is set to 0.1, and \textit{max-step} is set to 100. 

In steganography, bpp is the metric used to measure the payload of  stego images,
$
\mathrm{bpp}=\frac{len(\mathbf{Secret})}{H \times W }.
\label{Eq.bpp}
$
Here, bpp means the number of secret bits carried by each pixel (bits per pixel), $len(\cdot)$ means the length of hidden secret data, \textit{H/W} is the height/width of stego images. 
Acc is used to measure the extraction accuracy of secret data:
$
\mathrm{Acc}=\frac{\mathbf{Secret^{*}} \bigodot \mathbf{Secret}}{\mathrm{len}(\mathbf{Secret})},
$
here, $\mathbf{Secret^*}$ and $\mathbf{Secret}$  mean the extracted and input secret data, respectively. $ \bigodot $ denotes the $XNOR$ operation. The security of steganography is evaluate by metric PE, which is defined as:
$
\mathrm{PE}=\min _{P_{F A}} \frac{1}{2}\left(P_{F A}+P_{M D}\right),
$
here, $ P_{FA} $ and $ P_{MD} $ are the false alarm rate and miss detection rate of stego images. PE ranges in $ [0,  1.0] $, and the optimal value of PE is 0.5. At this time, the steganalyzer cannot distinguish the source of images and can only guess at random.  A pre-trained Resnet50 functions as the quality assessor. It is trained on 20k real images of CelebA and 20k generated images of Glow.  The images generated by the original un-modified \textbf{Z} are called plain images, in which no secret data is hidden. The images generated by the stego latent $ \mathbf{Z_s} $ are called stego images, which contain secret data.

\subsection{The effect of latent optimization strategy}
\begin{figure}[t]
	\centering
	\includegraphics[width=0.9\linewidth]{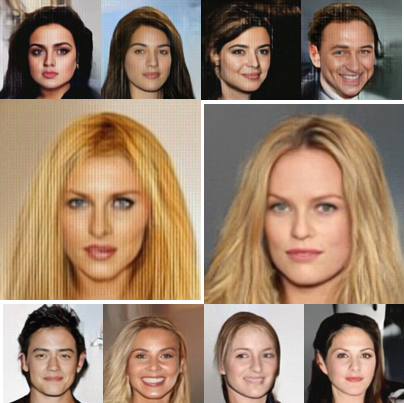} 
	\caption{result images (top + middle-left / bottom + middle-right) before and after using the latent optimization strategy.}
	\label{Fig.img_com}
\end{figure}
\begin{figure}[]
	\centering
	\includegraphics[width=0.9\linewidth]{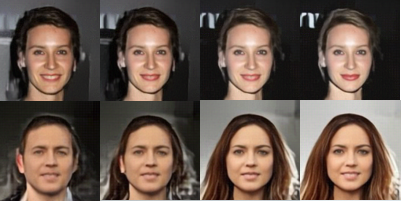} 
	\caption{images change slowly during latent optimizing.}
	\label{Fig.img_imp}
\end{figure}
The proposed latent optimization strategy is helpful in improving image quality. Fig.\ref{Fig.img_com} compares the generated images before/after using  the proposed strategy. The top row are the original generated images, which are of low fidelity with some white lines, mosaics and noises in the contents. The bottom row are the generated images after employing the proposed optimization strategy. The middle row shows the detailed images. Compared with the un-optimized images, we can find that our proposed optimization strategy can increase the image quality greatly. 

Meanwhile, the image content may be changed gradually during the optimization process. It can be observed from Fig.\ref{Fig.img_imp}, facial expression, hairstyle and gender are changing smoothly. At the same time, these images are getting clearer.

\subsection{The performance of proposed GSF}
Depending on whether the stego images are saved with rounding operation, we develop two strategies to generate stego images. Among them, secret data is hidden in different bit positions of encoded binary sequences of latent \textbf{Z}.

\begin{table}[b]
	\centering
	\small
	\setlength{\tabcolsep}{2.3mm}{
		\caption{ our scheme's performance for PNG stego images.}
		\label{tab-png}
		\begin{tabular}{@{}c|ccccc@{}}
			\toprule
			0 bpp &3 bpp	&6 bpp	&9 bpp &36 bpp	&72 bpp \\ 
			\midrule
			$[S, \alpha:\beta]$ & \textit{S} &\textit{S}, 22:22	&\textit{S}, 21:22 &\textit{S}, 12:22	&\textit{S}, 0:22 \\ 
			\midrule
			~~ Acc (\%)	&92.44	&74.09	&64.65 &54.19	&51.84         \\ 
			\midrule
			\multicolumn{5}{@{}l@{}}{\begin{minipage}{0.45\linewidth} \includegraphics[width=3.25in]{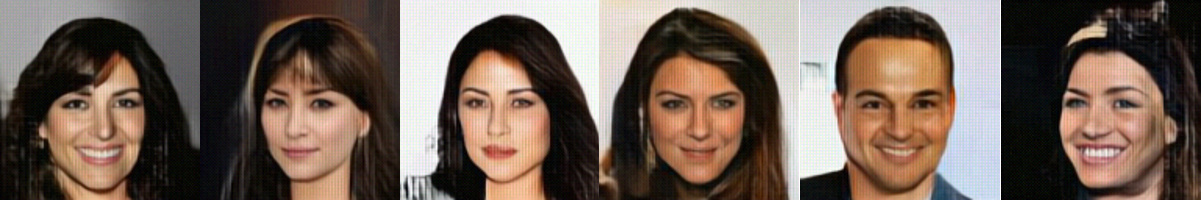} \end{minipage}} \\														            
			\bottomrule   						
	\end{tabular}}
\end{table}

In the first case,  the generated stego images are rounded to integers between [0, 255] and saved to image formats such as PNG and uncompressed JPEG. Because the rounding operation is irreversible and a large number of pixels will be clipped during image saving, the extraction accuracy of secret data will be decreased. As shown in Table \ref{tab-png}, 
the leftmost column are the generated plain images (0 bpp, images without secret data), which are generated with the unmodified  latents. The other columns are the performances of generated stego images with various payloads.  $[S, \alpha : \beta]$  means the $Sign $ bit and the bits from $\alpha_{th}$ to $\beta_{th}$ used for data hiding.  Different  payloads can be acquired by changing the value of $\alpha$ with $\beta$ fixed to 22.  Acc is the average extraction accuracy of secret data. The Acc value decreases slowly as the increase of payload, for the bits with smaller serial numbers become more difficult to be extracted after pixel clipping and rounding operation. The maximal payload of our scheme is 72 bpp with the Acc value of 51.84\%.  When the payload is 3 bpp,  Acc obtains a relatively high value of 92.44\%. In fact, it can reach 100\% if an error correction code is used.  The generated stego images look totally different from the plain images, because modifying the $ Sign $ bits will invert the pixel values and cause great influence to image content. However, the generated stego images are of low quality in general, which may arouse suspicion of the third-party.

%\begin{figure}[]
%	\centering
%	\includegraphics[width=0.95\linewidth]{fig/sign.png} 
%	\caption{comparison of generated PNG plain/stego  images. The top row is the originally generated plain images of Glow with the unmodified latent, and the bottom row is the stego images when the $ Sign $ bit of binarized sequences  are concealed with random secret data (size: $128 \times 128 $, payload: 3 bpp, mean Acc: 91.33\%).}
%	\label{Fig.sign}
%\end{figure}

%\begin{table*}[ht]
%	\centering
%	\renewcommand\arraystretch{1.2}
%	\setlength{\tabcolsep}{4.2mm}{
%		\caption{the performances of our proposed GSF.}
%		\label{Tab.perf}
%		\begin{tabular}{@{}c|cccccccc@{}}
%			\toprule
%			0 bpp &3 bpp	&6 bpp	&12 bpp	&18 bpp	&27 bpp	&30 bpp	&48 bpp	&69 bpp \\ 
%			\midrule
%			~[$\alpha$:$\beta$]	&[22: 22]	&[21: 22]	&[19: 22]	&[17: 22]	&[14: 22]	&[13:22]	&[7: 22]	&[0: 22] \\
%			\midrule
%			~~ Acc (\%)	&99.95	&99.93	&99.82	&99.59	&98.53	&97.76	&85.61	&74.71         \\ 
%			\midrule
%			\multicolumn{9}{@{}c@{}}{\begin{minipage}{0.98\linewidth} \includegraphics[width=6.85in, trim=0 0 0 52, clip]{fig/F8-1.png} \end{minipage}} 															\\            
%			\midrule
%			~~ Acc (\%)	&99.98	&99.97	&99.92	&99.81	&99.26	&98.80	&89.47	&77.32          \\ 
%			\midrule
%			\multicolumn{9}{@{}c@{}}{\begin{minipage}{0.98\linewidth} \includegraphics[width=6.85in, trim=0 0 0 52, clip]{fig/F8-3.png} \end{minipage}} 															\\                                                         
%			\bottomrule   						
%	\end{tabular}}
%\end{table*}

\begin{table}[]
	\centering
	\small
	\setlength{\tabcolsep}{2.95mm}{
		\caption{scheme's performance for TIFF stego images.}
		\label{Tab.perf}
		\begin{tabular}{@{}c|ccccc@{}}
			\toprule
			0 bpp  & 3 bpp &3 bpp	&27 bpp	&48 bpp	&69 bpp  \\ 
			\midrule
			$[S, \alpha:\beta]$ & \textit{S}	&22:22	&14:22	&7:22	&0:22 \\
			\midrule
			~~ Acc (\%)	&99.99 &99.95	&98.53	&85.61	&74.71        \\ 
			\midrule
			\multicolumn{5}{@{}l@{}}{\begin{minipage}{0.48\linewidth} \includegraphics[width=3.25in]{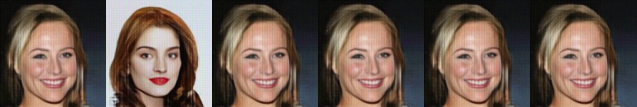} \end{minipage}} 															\\            
			\midrule
			~~ Acc (\%) &100	&99.98	&99.26	&89.47	&77.32          \\ 
			\midrule
			\multicolumn{5}{@{}l@{}}{\begin{minipage}{0.48\linewidth} \includegraphics[width=3.25in]{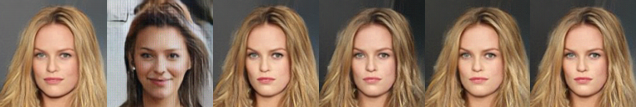} \end{minipage}} 															\\                                                         
			\bottomrule   						
	\end{tabular}}
\end{table}

In the second case, the generated stego images are saved with float datatype without rounding operation, i.e., images can be saved as TIFF format with the datatype of float32. Thus,  no information loss is introduced in saving. Actually, floating TIFF format is widely used in the scenes that require high precision, such as aerial images.  We adopt the proposed latent optimization strategy here to improve the quality of TIFF stego images. In the table, both the random latent ($ \mathbf{Z} \sim N (0, 1) \times \delta $) and optimized latent (\textbf{Z} after  optimization) are used to generate TIFF stego images, and the results are listed in Table \ref{Tab.perf}.  We can see the proposed latent optimization strategy can generate higher quality stego images, however, modifying the $ Sign$ (\textit{S}) bit will deteriorate the image quality. To maintain the fidelity of stego images, we only modify the fractional $\alpha_{th}-\beta_{th}$ bits of binarized sequences to hide secret data. In the table,  Acc decreases gradually as payload increases. Our GSF achieves a maximum payload of 69 bpp with an Acc value of 77.32\%. When the payload is lower than 27 bpp, the Acc values are above 99\%. 

\begin{figure}[b]
	\centering
	\subfigure[3 bpp PNG stego images]{
		\includegraphics[width=0.45\linewidth]{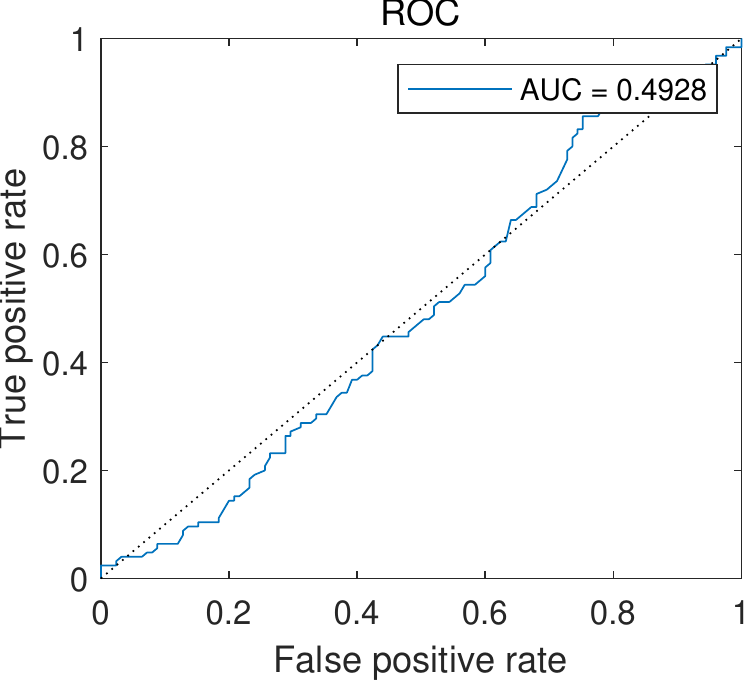}
		%\caption{fig1}
	}
	\subfigure[27 bpp TIFF stego images]{
		\includegraphics[width=0.45\linewidth]{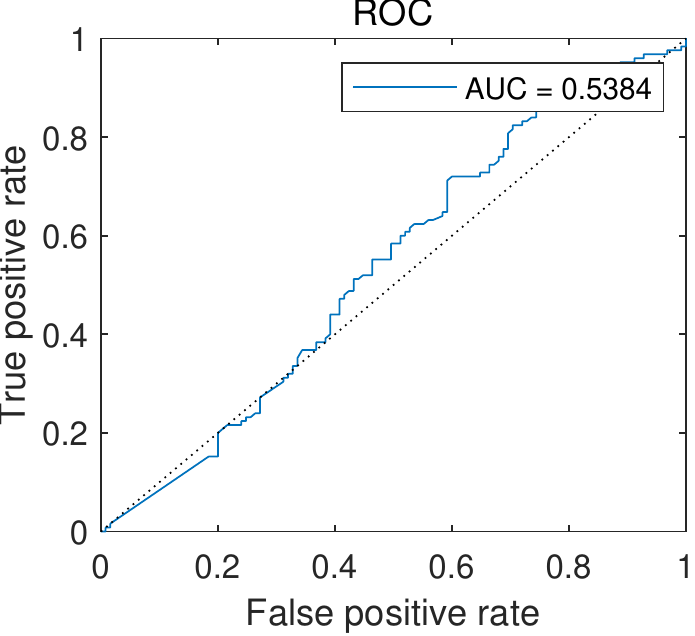}
	}
	\caption{ ROC curves of our stego images tested on SCRMQ1.}
	\label{Fig.roc}
\end{figure}

\begin{table}[]
	\centering
	\small
	\setlength{\tabcolsep}{1.8mm}
	\renewcommand\arraystretch{1.3}
	\caption{PE values of generated stego images}
	\label{Tab.PE}
	\begin{tabular}{cccccc} 
		\toprule
		Image format         &  bpp & SCRMQ1 & Ye-net & SR-net   & Siamese \\ 
		\midrule
		PNG                 &  3                 & 0.4780       & 0.500       &  0.493                  & 0.473           \\ \midrule
		\multirow{2}{*}{TIFF} & 18        & 0.4998 & 0.458 & 0.470 & 0.304        \\
		& 27                   & 0.4824 & 0.416 & 0.404 & 0.252        \\ 
		\bottomrule
	\end{tabular}
\end{table}

The security of our proposed GSF is tested with four advanced steganalysis methods: SCRMQ1\cite{color_rich}, Ye-net\cite{yenet}, SR-net\cite{srnet} and Siamese-CNN\cite{siamese}. SCRMQ1 is a traditional feature extractor for  colored stego images, the results of which are classified by an ensemble classifier. Ye-net, SR-net and Siamese-CNN are three SOTA DL-based steganalysis algorithms. In detection, the generated plain images (0 bpp) are used as the cover dataset, and the generated stego images are used as the stego dataset. The ROC curves of our generated stego images are as plotted in Fig.\ref{Fig.roc}. They both get an AUC value close to 0.5 (the optimal value). The detailed detection PE values are listed in Table \ref{Tab.PE}. As we can see, most of the PE values are close to ideal 0.5, which indicates our method is very safe.  It means existing steganalysis methods can hardly distinguish our generated stego images from cover images.

\subsection{Results Comparison}
\begin{figure}[]
	\centering
	\includegraphics[width=0.95\linewidth]{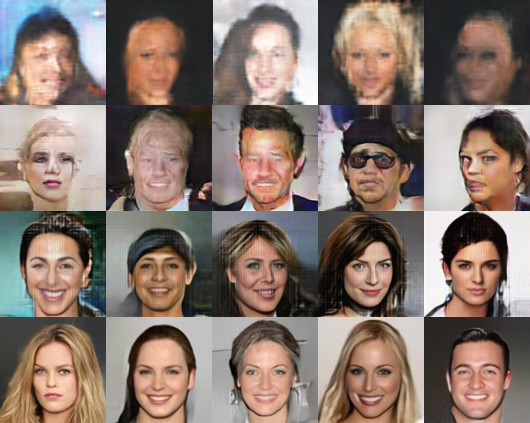} 
	\caption{Visual comparison of generated stego images. The top two rows are the stego images generated by Zhang\cite{zhang2020generative} (payload: 5 bits/image) and Hu\cite{hu2018novel} (300 bits/image); the bottom two rows are the $128\times128$ PNG and TIFF stego images generated by our GSF, with the payloads of 3 bpp (49,152 bits/image) and 24 bpp (393,216 bits/image), respectively.} 
	\label{Fig.Hu}
\end{figure}

We compare our GSF with SOTA works as listed in Table \ref{Tab.comp}.  Zhang\cite{zhang2020generative} and Hu\cite{hu2018novel} are two SOTA GS networks based on GANs, which use the semantic labels and noise vectors of GANs to hide secret data. We re-implemented two works on dataset CelebA with the default settings and net structures. We compare our scheme with the best results of them. It should be notice that the Acc values for Zhang and Hu are obtained from float-point TIFF images.  Fig.\ref{Fig.Hu} demonstrates the generated stego images of different methods. To evaluate the image quality, we use a non-reference assessor NIQE\cite{niqe} in our scheme. Where, lower NIQE score means better image quality. The security of all the algorithms is evaluated with metric PE by steganalyzer Ye-net\cite{yenet}. In  detection, we use the generated plain images (containing no secret data) and stego images (containing random binary secret data) as the cover and stego datasets. From the results, we can see our proposed GSF  performs much better than the other two works in almost all metrics. The hiding capacity of our GSF is more than 40 times than that of the other methods. Meanwhile, higher secret extraction accuracy, better stego image quality and superior security level are also obtained by our scheme.

\begin{table}[htb]
	\centering
	\caption{comparison of proposed GSF with SOTA methods.}
	\label{Tab.comp}
	\small
	\setlength{\tabcolsep}{2.8 mm}
	\renewcommand\arraystretch{1.3}
	\begin{tabular}{@{}lcccc@{}}
		\toprule
		methods                         & bpp $\uparrow$         & Acc (\%) $\uparrow$   &NIQE $\downarrow$   & PE $\rightarrow 0.5 $            \\ 
		\hline
		Zhang(\cite{zhang2020generative})										& 1.22e-3	& 71.85		&56.60	&0.48			\\
		Hu (\cite{hu2018novel})                             & 7.32e-2    & 90.50      &14.13     & 0.49             \\ 
		\textbf{GSF\_PNG }                         & \textbf{3} &   \textbf{92.44}   & \textbf{14.01}  & \textbf{0.50} \\ 
		\textbf{GSF\_TIFF}                          & \textbf{24} & \textbf{99.53} & \textbf{12.96}  & 0.44 \\ \bottomrule
	\end{tabular}
\end{table}

\section{Conclusion}
A novel flow-based generative steganography solution is proposed in our research, which can generate realistic stego images and recover the hidden secret with a single network. We first build a reversible bijective mapping to convert secret data to stego images, and recover the hidden secrets exactly by reversing the mapping. Furthermore, a novel latent optimization strategy is developed to improve the fidelity of generated stego images. Compared with existing works, our proposed GSF has better performances with higher hiding capacity, extraction accuracy,  image quality and security.

\noindent \textbf{Acknowledgement:} This work was supported by the National Natural Science Foundation of China U20B2051, 62072114, U20A20178, U1936214.
\bibliographystyle{IEEEbib}
\bibliography{ref}

\end{document}